\documentclass[letterpaper, 10 pt, conference]{ieeeconf}  

\IEEEoverridecommandlockouts                              

\usepackage{graphics} 
\usepackage{graphicx} 
\usepackage{multirow} 
\usepackage{paralist} 
\usepackage{amsmath} 
\usepackage{amssymb}  

\title{\LARGE \bf
Unsupervised Driving Event Discovery Based on Vehicle CAN-data 
}

\author{Thomas Kreutz$^{1}$, Ousama Esbel$^{2}$, Max Mühlhäuser$^{1}$ and Alejandro Sanchez Guinea$^{1}$
\thanks{$^{1}$The authors are with the Telecooperation Lab at the Technical University Darmstadt, Germany {\tt\small \{kreutz, max, sanchez\}@tk.tu-darmstadt.de}}%
\thanks{$^{2}$The authors are with COMPREDICT GmbH, Darmstadt, Germany {\tt\small \{esbel\}@compredict.de}}
}

\begin{document}

\maketitle
\thispagestyle{empty}
\pagestyle{empty}

\begin{abstract}



The data collected from a vehicle's Controller Area Network (CAN) can quickly exceed human analysis or annotation capabilities when considering fleets of vehicles, which stresses the importance of unsupervised machine learning methods. This work presents a simultaneous clustering and segmentation approach for vehicle CAN-data that identifies common driving events in an unsupervised manner. The approach builds on self-supervised learning (SSL) for multivariate time series to distinguish different driving events in the learned latent space. We evaluate our approach with a dataset of real Tesla~Model~3 vehicle CAN-data and a two-hour driving session that we annotated with different driving events. With our approach, we evaluate the applicability of recent time series-related contrastive and generative SSL techniques to learn representations that distinguish driving events. Compared to state-of-the-art (SOTA) generative SSL methods for driving event discovery, we find that contrastive learning approaches reach similar performance.


\end{abstract}

\section{Introduction}
Deep learning is being applied successfully in the automotive industry to many use cases such as autonomous driving~\cite{luckow2018artificial} or predictive maintenance (PdM)~\cite{foulard2017fahrzeugkomponenten}. 
For instance, for PdM, multivariate time series data collected from a vehicle’s Controller Area Network (CAN) can be used to estimate the health condition of a car component~\cite{foulard2017fahrzeugkomponenten}. Furthermore, as CAN-data reflects the overall state of the vehicle, based on this data, it is possible to infer information about the current driving event, situation, or behavior~(e.g., \cite{hallac2018drive2vec, bender2015unsupervised}). 

A car can generate multiple gigabytes of vehicle CAN-data in a single driving session~\cite{singh2019deep}, which may grow quickly to multiple tera- or petabytes when considering fleets of cars. For this reason, it is a time-consuming and expensive task for domain experts to manually annotate or analyze these enormous amounts of data~\cite{bender2015unsupervised}. Consequently, problems like the sampling bias may limit the performance of, for instance, PdM models trained on non-reviewed data. A possible solution is using unsupervised machine learning to discover information in extensive unlabelled collections of data. Recent work has used self-supervised learning (SSL) to transform unlabelled vehicle CAN-data into a new representation that allows distinguishing different driving events or behavior~(e.g., \cite{hallac2018drive2vec, liu2017visualization}) without any prior knowledge.

\begin{figure}[t]
    \centering
    \includegraphics[width=\linewidth]{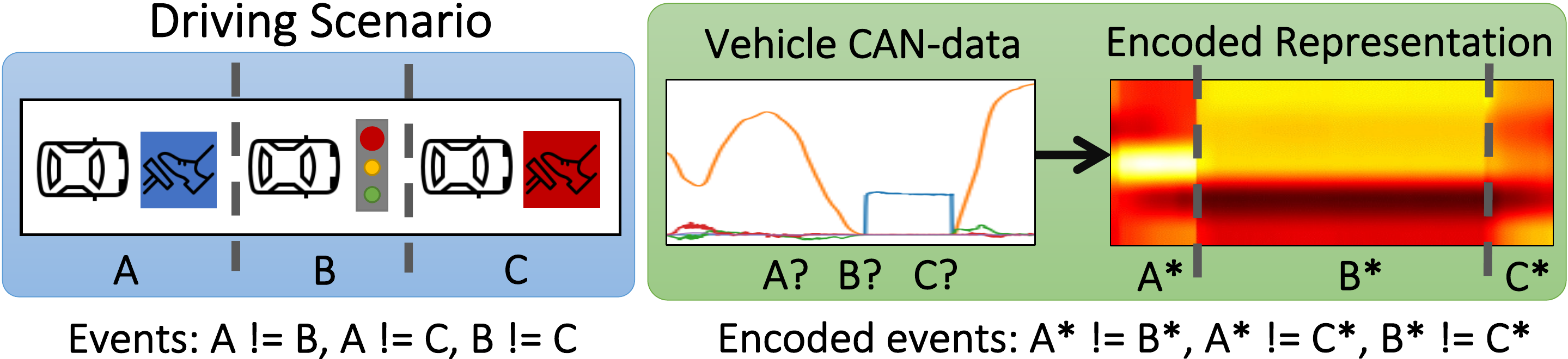}
    \caption{A neural network encodes vehicle CAN-data into a representation, where the same driving events become similar at all time steps. Such a representation allows to cluster each time step in isolation, which simultaneously clusters and segments vehicle CAN-data into driving events.
    }
    \label{fig:problem}
\end{figure}

This paper focuses on the unsupervised discovery of driving events by proposing a method that simultaneously clusters and segments vehicle CAN-data. Consider the driving events A, B, and C composing a driving scenario in Fig.~\ref{fig:problem}. State-of-the-art (SOTA) SSL methods for vehicle CAN-data (e.g. Drive2Vec~\cite{hallac2018drive2vec}) learn representations of vehicle CAN-data, where all time steps that are part of, e.g., event type $A$ share a similar representation $A^{*}$, and all time steps of other events are different ($B^{*},C^{*}$). Due to this property in the encoded representation, we can cluster all individual time steps in isolation and obtain a state sequence that is 1) constant for the duration of events and 2) only differs at the event changepoints, which effectively segments and clusters the data the same time.

Previous work on unsupervised driving event discovery can be classified into motif discovery~(e.g.,~\cite{silva2021tripmd}), time series segmentation~(e.g.,~\cite{takenaka2012contextual, bender2015unsupervised}), subsequence clustering (e.g.,~\cite{shouno2018deep, siami2020mobile}), 
simultaneous segmentation and clustering~(e.g.,~\cite{hallac2017toeplitz}) and generative SSL methods for representation learning (e.g.,~\cite{liu2017visualization, hallac2018drive2vec}). Aside from the latter SSL methods and the approaches in~(\cite{shouno2018deep, hallac2017toeplitz}), the majority of previous methods first uses a segmentation or motif discovery algorithm to discover potential event segments, extracts segment level features with, e.g., a neural network, and then clusters the resulting segments in a mandatory final step.

Similar to and inspired by Toeplitz Inverse Covariance-based Clustering (TICC)~\cite{hallac2017toeplitz} and Variational Animal Motion Embedding (VAME)~\cite{luxem2020identifying}, our approach simultaneously clusters and segments vehicle CAN-data. Because our approach produces a sequence of event states that are split into segments at their changepoints, it is not limited to a maximum length of events as in~\cite{shouno2018deep} and is independent of segments obtained by a segmentation or motif discovery algorithm. 

Furthermore, we evaluate recent contrastive SSL methods for representation learning of vehicle CAN-data. To the best of our knowledge, we are the first to explore contrastive SSL methods for this kind of data. Recent advances in deep learning-based time series changepoint detection (CPD)~\cite{deldari2021time} and contrastive SSL for multivariate time series~(e.g.,~\cite{franceschi2019unsupervised, tonekaboni2021unsupervised}) motivate to research the applicability of contrastive SSL for vehicle CAN-data. Contrastive SSL has shown extraordinary results in distinguishing different states when applied to human activity recognition (HAR) data~\cite{deldari2021time}, electroencephalography (EEG) data~\cite{tonekaboni2021unsupervised}, or household consumption data~\cite{franceschi2019unsupervised}, among others. 
The main contributions of this work are as follows:
\begin{compactitem}
    
    \item 
    An approach to discover driving events in an unsupervised manner, which simultaneously clusters and segments vehicle CAN-data. 
   
    \item An empirical evaluation of contrastive SSL methods for vehicle CAN-data, which is (to the best of our knowledge) the first step in the direction of applying these kinds of methods to vehicle CAN-data.
    
\end{compactitem}

\section{Notations and Definitions}

We define a multivariate time series $MTS$ as a tuple $MTS := (X_{1}, X_{2}, ..., X_{N})\in \mathbb{R}^{d \times N}$
where each element $X_{t} \in MTS$ represents a $d$-dimensional measurement such that $X_{t} = (x_{1}, ..., x_{d})$ for all $0 < t \leq N$.

We define a subsequence $S$ of $MTS$ as a collection of points starting at a time point $k$ and ending at index $l$. Formally, we define this subsequence as 
\mbox{$S_{k,l} = (X_{k}, ..., X_{l}) \subseteq MTS$} where $\subseteq$ denotes that $S$ is a subsequence of $MTS$.

A sliding window is a fixed size window of length $w$ that moves sequentially over $MTS$. For an $MTS$ of length $N$ and for each time point $X_{t} \in MTS$ with $w \leq t \leq N$, $w << N$, we can define a sliding window as a subsequence of this $MTS$ with $S_{t-(w-1), t} \subseteq MTS$\footnote{Any timepoint $l=t$ is the end of a sliding window, and $k=t-(w-1)$ is the start. The first possible sliding window always starts at $k=1$ and ends at $t=w$ and the last sliding window always ends with $l=N$. This reduces the time series by $w-1$ time steps, and $w$ is the new start point.}. 

\section{Approach}

\begin{figure}[t]
    \centering
    \includegraphics[width=0.7\linewidth]{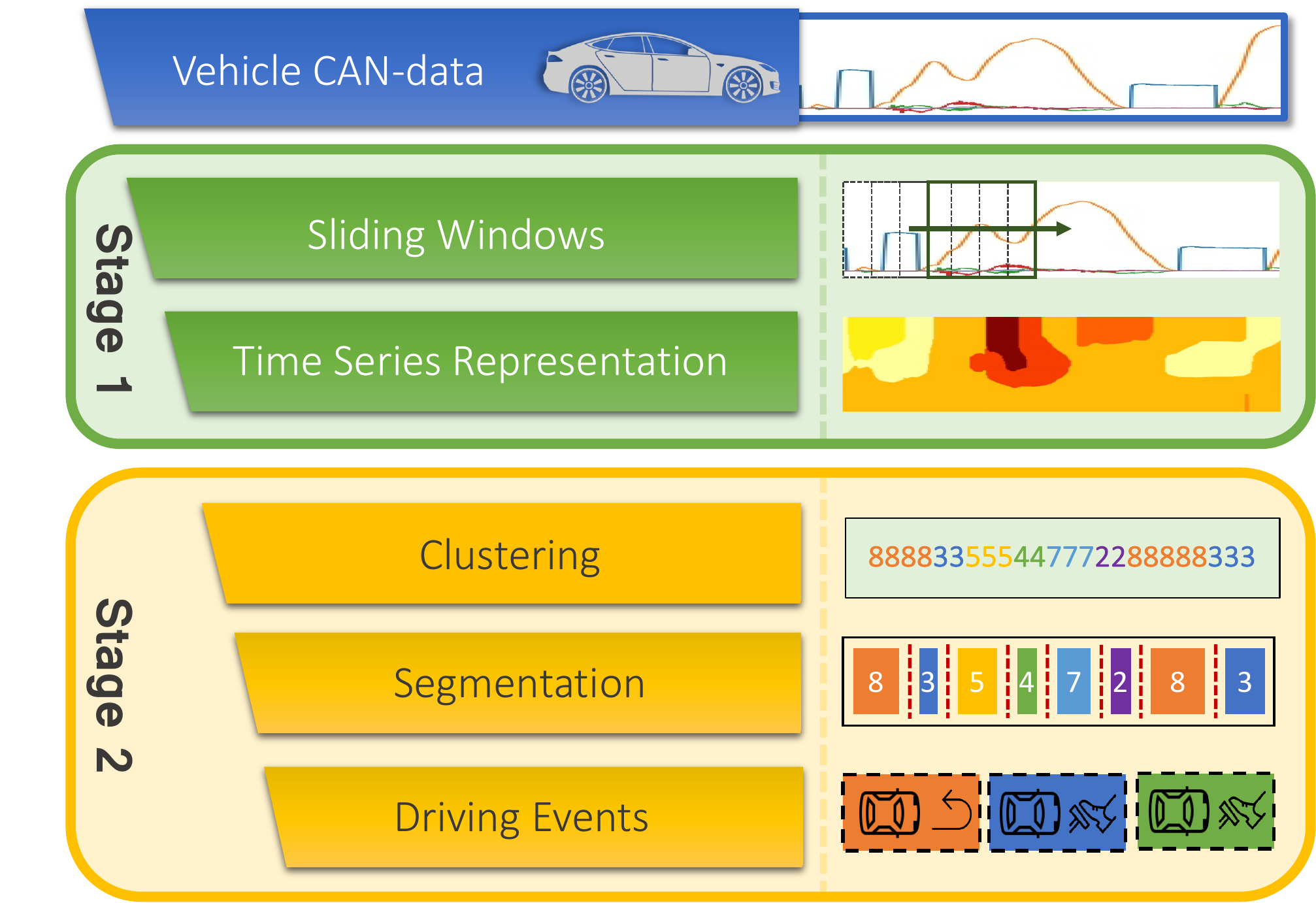}
    \caption{High level overview of the proposed two-stage approach}
    \label{fig:approach}
\end{figure}

A high-level overview of our two-stage approach is depicted in Fig.~\ref{fig:approach}. In Stage~1, we transform vehicle CAN-data into a new representation by using the sliding window technique and encoding each sliding window with a neural network. The encodings are concatenated to compose a new time series which is enriched at each time step with information about the current driving event state. Afterward, in Stage~2, we cluster the sliding window representations and obtain a one-dimensional cluster sequence, which can be considered a discrete driving event state sequence. The state sequence is segmented in variable-length driving events by splitting the sequence at the cluster changepoints. The corresponding constant cluster of a segment characterizes the respective driving event. 

\subsection{Stage 1 - Time Series Representation}
\begin{figure}[t]
    \centering
    \includegraphics[width=0.9\linewidth]{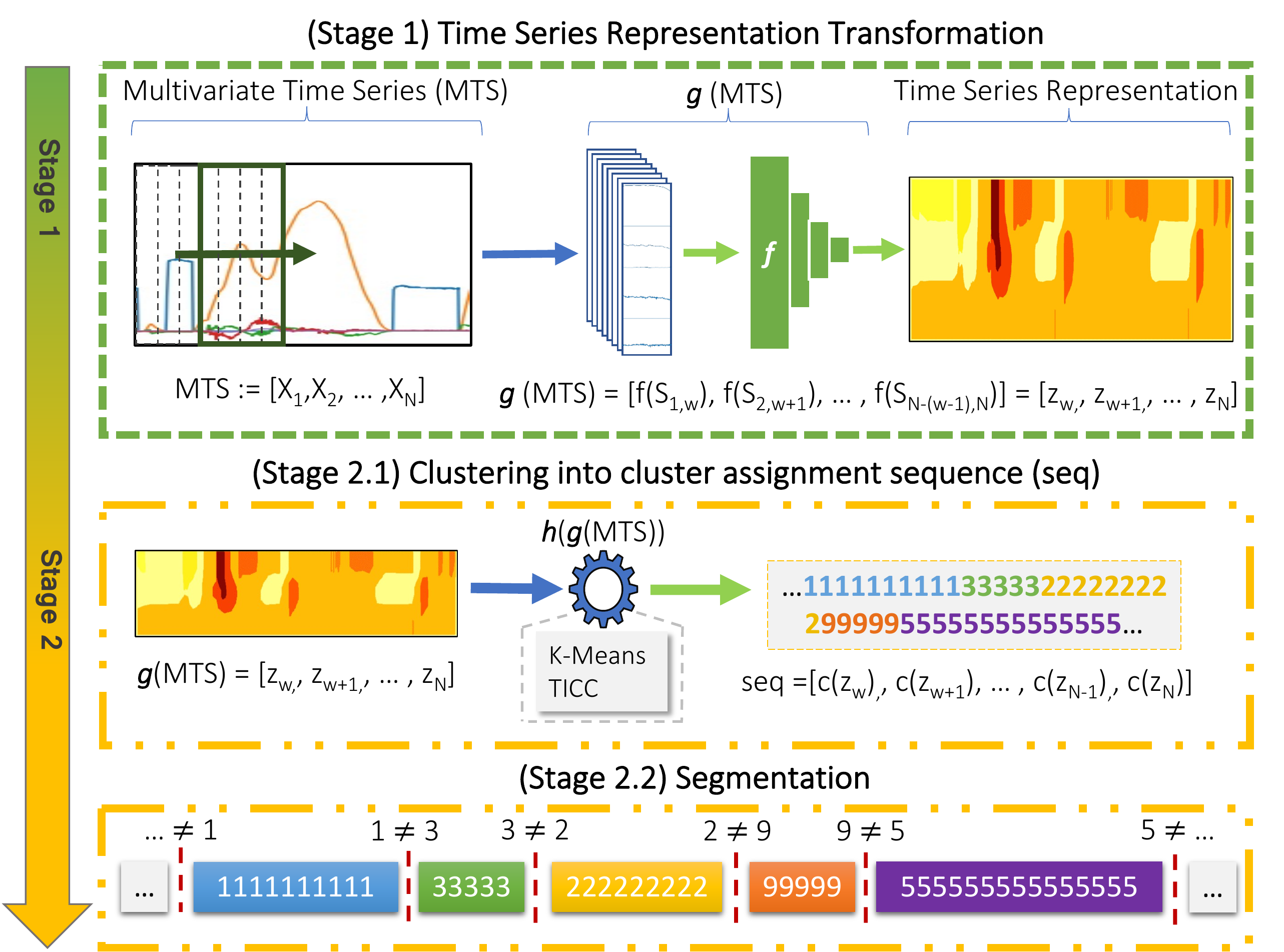}
    \caption{Detailed overview of both stages of the approach. Stage~1 concerns with the time series representation, Stage~2 concerns with the clustering and segmentation}
    \label{fig:approach_detailed_overview}
\end{figure}

We summarize our approach in more detail in Fig.~\ref{fig:approach_detailed_overview}. The \textit {time series representation} in Stage~1 is obtained by first sliding a fixed-size window of length $w$ over a $d$-dimensional multivariate time series $MTS \in \mathbb{R}^{ d \times N}$ of length $N$. Afterward, each window is encoded to a new representation. The resulting representations are concatenated to a new multivariate time series that has a different dimension but preserves the temporal order. 



We consider a representation to be a mapping from a $d \times w$-dimensional sliding window input to a new $e$-dimensional representation. This mapping is defined by an encoder function~$f : \mathbb{R}^{d \times w} \longrightarrow \mathbb{R}^{e}$. The function $f$ can be interpreted as the encoder part of an Autoencoder (AE). For $f$, we use the concept of a \textit{timestamp-level representation}, where each point in time $X_{t} \in MTS$ with $w \leq t \leq N$ is mapped to a new representation $z_{t}$. We refer to the representation $z_{t} = f(S_{t-(w-1),t})$ as a \textit{timestamp-level representation}, which is the encoding of a sliding window $S_{t-(w-1),t}$. 

We define a \textit {time series representation} as an $e$-dimensional representation of a $d$-dimensional $MTS$, where each sliding window was encoded with $f$. Formally, the \textit {time series representation} is a function \mbox{$g: \mathbb{R}^{d \times N} \longrightarrow \mathbb{R}^{e \times N-(w-1)}$} that maps (encodes) the elements of $MTS$ to an $e$-dimensional representation and preserves their temporal order. By applying $f$ to all sliding windows over $MTS$, we obtain the \textit {time series representation} function $g$, defined as:
\begin{align} \label{eq:timeseries_representation_transformation_function}
    g(MTS) &= [f(S_{1, w}), f(S_{2, w+1}), ... , f(S_{N-(w-1), N})] \\
           &= [z_{w}, z_{w+1}, ..., z_{N}]
\end{align}

\subsection{Stage 2 - Simultaneous Segmentation and Clustering}
A detailed overview of Stage~2 is depicted in Fig.~\ref{fig:approach_detailed_overview}. In Stage~2.1, each \textit{timestamp-level representation} \mbox{$z \in [ z_{w}, z_{w+1}, ..., z_{N} ] = g(MTS)$} is clustered, with the number of clusters fixed to $k$. The clustering process can be formally defined by a partition $h := \{ H_{1}, ... , H_{k} \}$ that partitions a \textit{time series representation} into $k$ disjoint non-empty subsets. The partition $h$ is produced by a clustering algorithm (e.g., \textit{k}-means or TICC) with $|h| = k$. Further, we denote the assignment of a \textit{timestamp-level representation} $z$ to its cluster by a function $c: \mathbb{R}^{e} \longrightarrow \mathbb{N}$. For any $z \in \mathbb{R}^{e}$, $c(z)$ returns the index $j$ of the cluster where $z \in H_{j}$. 
We apply $c$ to each $z \in [z_{w}, ..., z_{N}]$ and obtain a sequence of cluster assignments $seq = [c(z_{w}), ..., c(z_{N})]$. 

\begin{equation}
    \forall z \in \mathbb{R}^{e}. \ c(z) = j \implies z \in H_{j} 
\end{equation}

In Stage~2.2., a simple segmentation step splits the cluster assignment sequence $seq$ at the changepoints into segments. Finally, each segment is assigned to its constant cluster assignment, which directly clusters the obtained segments.

\subsubsection*{Driving Event Discovery}
The discovery of driving events from the segmentation is achieved by grouping all segments under their corresponding constant cluster assignment. The segments of a cluster can be interpreted by, e.g., a domain expert, or they can be compared to labeled data to find high overlaps between a cluster and ground truth annotations. Finally, each segment in the time series can be annotated with the driving event interpretation of its cluster. The segmentation can be visualized, for example, on a GPS trajectory by coloring each time step with its cluster assignment.

\subsection{Application to Multiple Multivariate Time Series}
Our approach can easily scale to multiple $MTS$ when considering Stage~1 with a collection of multivariate time series ($C_{MTS}$) instead of a single $MTS$. In this setting, Stage~1 is carried out for each $MTS \in C_{MTS}$ individually and we obtain a collection of time series representations \mbox{$C_{z} := \{ \ g(mts) \  | \ mts \in C_{MTS}\ \}$}. In Stage~2.1, the partition $h$ is computed on all datapoints in $C_{z}$. Finally, we can obtain a collection of cluster assignment sequences \mbox{$C_{seq} := \{ \ [c(z_{a}), ..., c(z_{b})] \ | \ [z_{a}, ..., z_{b}] \in C_{z} \ \}$}.

\subsection{Neural Network Encoder Training}

\begin{figure}[t]
    \centering
    \includegraphics[width=0.8\linewidth]{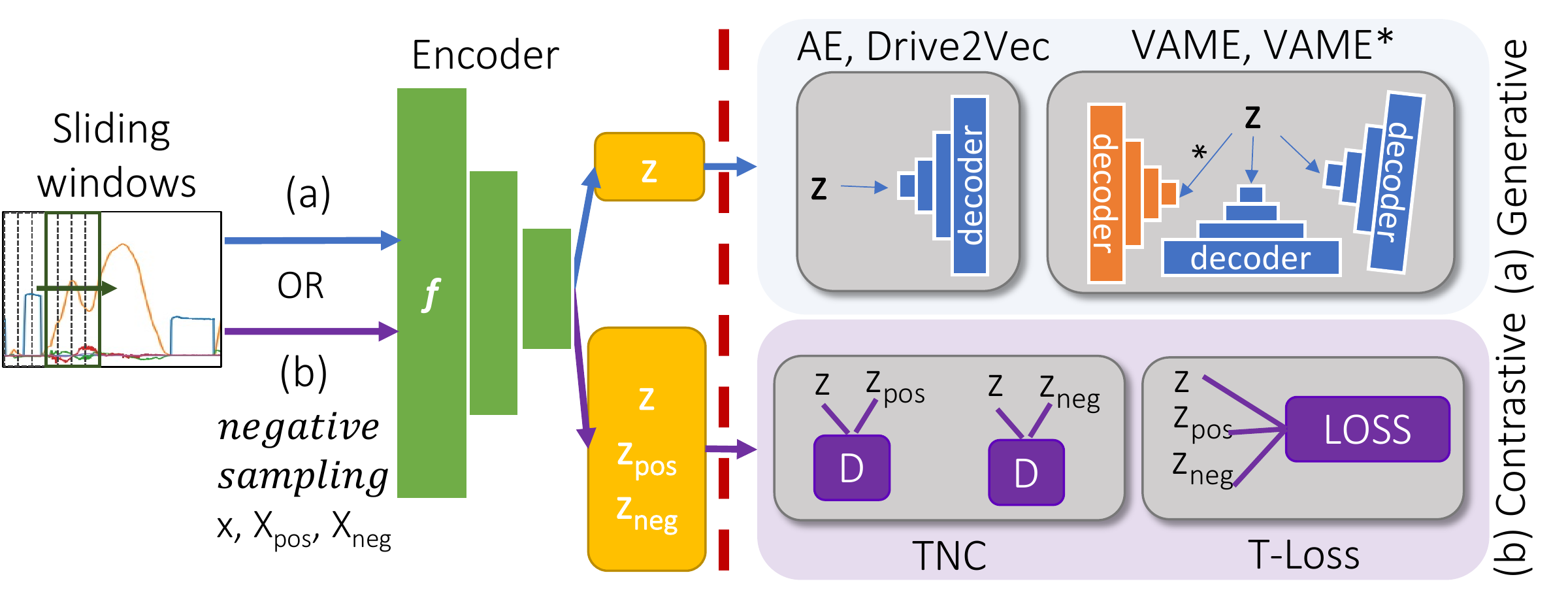}
    \caption{Overview of the neural networks used in this work. a) encoder-decoder models, b) contrastive learning models}
    \label{fig:neural_networks}
\end{figure}

Following the SSL taxonomy in~\cite{liu2021self}, the encoder $f$ can be trained in a generative or a contrastive manner. 
A generative (Encoder-Decoder) model is usually trained with a decoder~$f_{dec}$ that has the goal to reconstruct the $d$-dimensional input from the e-dimensional representation of~$f$ with minimal error. Contrastive learning directly trains the encoder with a contrastive loss that pushes similar data points to the same representation and non-similar data points to a different representation. 

As summarized in Fig.~\ref{fig:neural_networks}, as generative methods, we use a normal AE as a baseline that was used in~\cite{sama2020extracting}, and a Variational AE (VAE) with an additional future decoder and \textit{k}-means objective, i.e., VAME~\cite{luxem2020identifying}.  Furthermore, we compare these methods to a SOTA Encoder-Decoder model for vehicle CAN-data called Drive2Vec~\cite{hallac2018drive2vec}. Drive2Vec learns representations that are predictive about the future, helping to better differentiate the current driving situation. Similarly, VAME is trained with a present and a future decoder to learn representations that are predictive about the future. We extend VAME to VAME* with a past decoder to learn predictive representation about the past and the future. 

As contrastive learning methods, we consider Temporal Neighborhood Coding (TNC)~\cite{tonekaboni2021unsupervised} and a triplet loss approach (T-Loss)~\cite{franceschi2019unsupervised}. TNC assumes a time series to be stationary for the full duration of an event. Given a reference window, they propose estimating it's stationary temporal neighborhood with a statistical test. Afterward, random windows from this neighborhood are enforced to be encoded to a similar representation, and windows outside the neighborhood to a different one by using a discriminator on the embeddings. T-Loss accepts variable-length time series input and is trained with a triplet loss. Random subsequences of a reference window are enforced to be encoded similarly to it, but different to other random windows. Similar to TNC, a larger reference window can be seen as an event's temporal neighborhood.



More technical details about the selected SSL methods are out of scope in this manuscript and we must point to the original works for further reading. The SSL methods we consider can encode corresponding time steps similar to each other. Towards this end, TNC has been used for, e.g., HAR data~\cite{tonekaboni2021unsupervised}, T-Loss for Household Consumption data~\cite{franceschi2019unsupervised}, and an AE in~\cite{sama2020extracting} for CAN-data. Using future context for training has been shown to effectively encode events to similar feature representations by Drive2Vec in~\cite{hallac2018drive2vec} for CAN-data, and with VAME for mice behavior data in~\cite{luxem2020identifying}.  We focus on empirically evaluating how these methods compare as a basis for our approach when learning timestamp-level representations of CAN-data.



\subsection{Clustering Methods}
As shown in Fig.~\ref{fig:approach_detailed_overview}, we use the \textit{k}-means and TICC clustering algorithms to cluster the encoded time steps of vehicle CAN-data. We chose the \textit{k}-means algorithm as a method that does not include any temporal dependency between time steps in order to leverage the learned representations that distinguish different driving events. On the other hand, we chose TICC because it incorporates a temporal dependency between time steps by using a dependency network.

\section{Experimental Evaluation}
\begin{table*}[t]
    
    \caption{Results of the best performing parameter configurations for each model, G=Generative, C=Contrastive} 
    \centering
    \begin{tabular}{c|c|c|c|c|c|c|c|c|c|c|c|c}
        & \multicolumn{2}{c}{VAME* (G)} &  \multicolumn{2}{c}{VAME (G)} & \multicolumn{2}{c}{Drive2Vec (G)} & \multicolumn{2}{c}{T-Loss (C)} & \multicolumn{2}{c}{TNC (C)} & \multicolumn{2}{c}{AE (G)} \\
        \hline
        & $k$-means & TICC & $k$-means & TICC & $k$-means & TICC & $k$-means & TICC & $k$-means & TICC & $k$-means & TICC \\
        \hline
        $F_{1} $ $(\uparrow)$ & \textbf{0.469} &  0.399 & 0.464 & 0.374 & 0.430 & 0.378 & 0.392 & 0.377 & 0.338 & 0.294 & 0.310 & 0.286  \\
        \hline
        \hline
        $e$ & 20 & 20 & 20 & 20 & 15 & 10 & 5 & 15 & 5 & 20 & 20 & 10 \\
        \hline
        $w$ & 10 & 5 & 15 & 10 & 10 & 5 & 10 & 15 & 16 & 10 & 5 & 10 \\
    \end{tabular}
    
    \label{tab:metric_results}
\end{table*}

\subsection{Experimental Setup}
We use the neural networks and TICC implementations available in their respective GitHub repositories (TNC\footnote{https://github.com/sanatonek/TNC\_representation\_learning}, T-Loss\footnote{https://github.com/White-Link/UnsupervisedScalableRepresentation \\ LearningTimeSeries},  VAME\footnote{https://github.com/LINCellularNeuroscience/VAME}, TICC\footnote{https://github.com/davidhallac/TICC}). As suggested in~\cite{yue2021learning}, we have modified the TNC model to use an encoder with dilated causal convolutions. For Drive2Vec, we use our implementation that, instead of gated recurrent units (GRU), uses a temporal convolutional network (TCN). We implemented a version of the AE from~\cite{sama2020extracting} in the PyTorch\footnote{https://pytorch.org/} deep learning framework. 

\subsection{Use Case and Training Dataset}
We consider a use case from the automotive domain, where the goal is to discover frequently occurring driving events in vehicle CAN-data in an unsupervised manner. We have collected a dataset that consists of 60 multivariate time series recordings of a Tesla~Model~3 vehicle. The data has been recorded in real driving sessions by using a datalogger. The datalogger is mounted into the car to directly access the CAN-bus over an external interface. It records the values of 113 vehicle signals upsampled at a frequency of 200hz.
However, from these 113 signals, we select only a subset of nine signals that we consider to be relevant for driving events: `Brake pressure front left/right'~[$bar$], `Electric motor torque'~[$Nm$], `Accelerator pedal position'~[$\%$], `Steering wheel angle'~[$^{\circ}$], `Velocity'~[$km/h$], `Longitudinal acceleration'~[$m/s^{2}$], `Lateral  acceleration'~[$m/s^{2}$], and `Yaw rate'~[$^{\circ}$/s].
To preprocess the data, we follow related work (e.g., \cite{hallac2018drive2vec, liu2017visualization, hallac2017toeplitz}) and resample the data from 200Hz to 10Hz and apply Z-normalization separately on each channel.

\subsection{Ground Truth - Manual Annotation of Testing Data}

We have manually annotated a driving trip to compare our approach's results against a ground truth. The trip was around two hours long (following the evaluation in~\cite{hallac2017toeplitz}), and designed to include different traffic scenarios that we believe to occur on urban, highway, or secondary roads. 
For the annotation we considered 13 different labels: `Standstill', `Acceleration', `Deceleration', `Turn Left', `Turn Right', `Braking', `Keep Velocity (KV)', `Fast KV', `Autonom KV', `S Turn', `Reverse', `Lane Change', and `Roundabout'.

\subsection{Implementation Details}

Regarding the neural networks, we follow~\cite{franceschi2019unsupervised} and do not perform any hyperparameter optimization on the network parameters. We evaluate our approach's general hyperparameters: The window size~$w$ and the embedding dimension~$e$. In particular, we evaluate \mbox{$w \in [5,10,15,20]$} and \mbox{$e \in [3,5,10,15,20]$}. Each model is trained for 200 epochs (800 steps for T-Loss) with a learning rate of $1e-4$. Sliding windows with a step size of $3$ are used for training, and a step size of $1$ is used at the clustering and testing step. T-Loss is trained with a reference window that is $3 \times w$.

Concerning the clustering algorithms, the number of clusters is fixed to $k=13$, which matches the number of ground truth labels. The \textit{k}-means algorithm is run 15 times with random restarts. For the TICC\footnote{
The original work~\cite{hallac2017toeplitz} suggested ten or more iterations. The results with three iterations are promising, and we consider running more iterations in the future. More iterations can significantly improve the result.} 
algorithm we use the parameters $window\_size= 10$, $\lambda=5-e3$, $beta = 400$, $max\_it = 3$, $thresholds = 2e-5$. 

\subsection{Best Performing Approach Configurations}

We summarize the results for the best performing models for both clustering algorithms in Table~\ref{tab:metric_results}, which also shows the corresponding parameters $e,w$ in the bottom rows. The row $F_{1}$ summarizes the macro $F_{1}$ score over all classes. 
On average, VAME* and VAME outperform the other models. Drive2Vec follows closely, and AE performs the worst. T-Loss and TNC fall short behind the generative models. We will see in the following sections that this performance difference is due to the `Turn Right' events. For all other driving events, the performance is similar. In addition, from the results of the bottom $e$ and $w$ rows in Table~\ref{tab:metric_results}, we can see that a higher embedding dimension of 10-20 and a window size between 5-15 give the best results.

\subsection{Latent Space}

\begin{figure*}[t]
    \centering
    \includegraphics[width=0.7\linewidth]{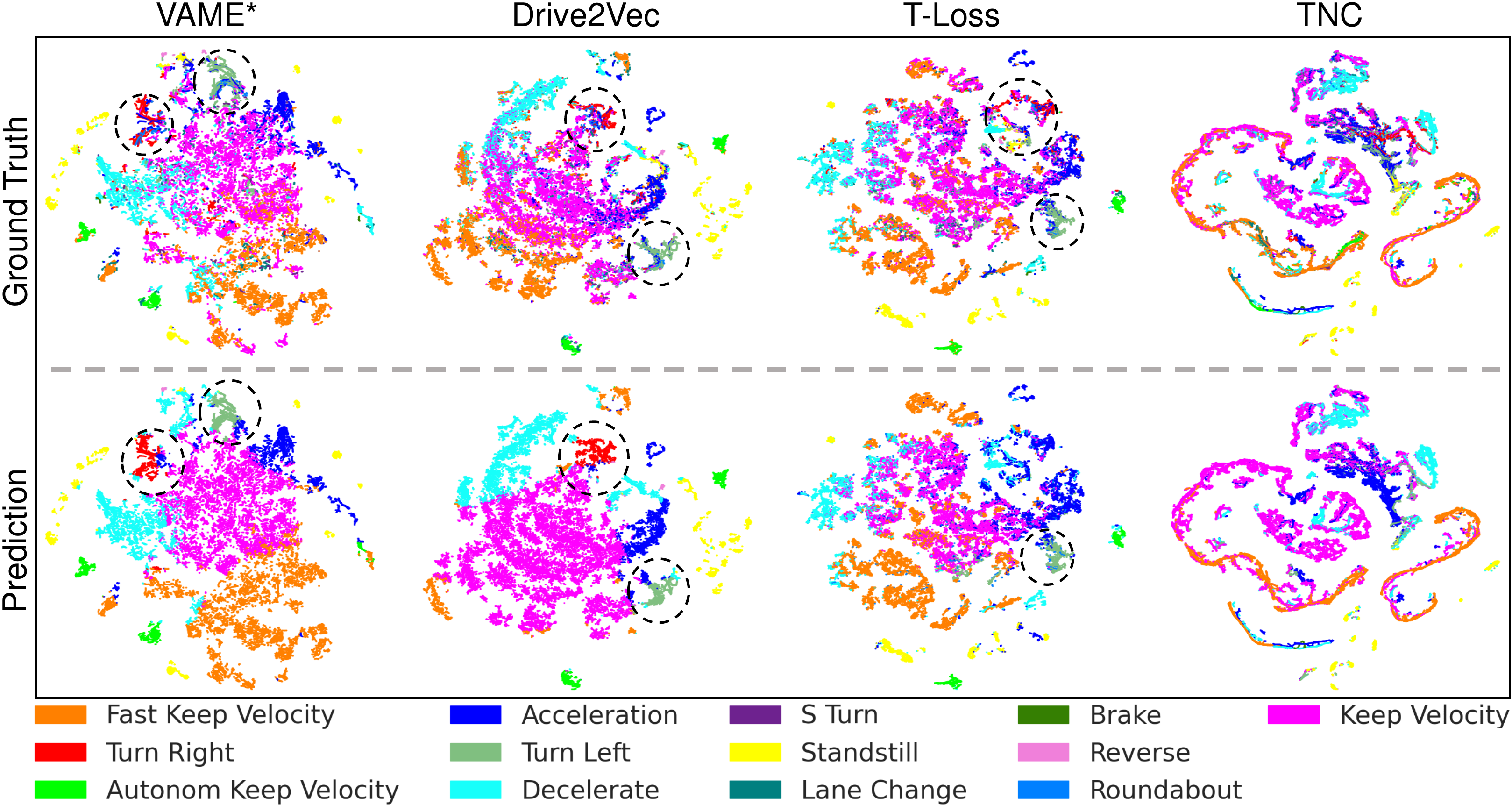}    \caption{Latent space visualization, where 20 dimensional representations of VAME* have been reduced to two dimensions with TSNE. Ground truth (top), and clustering (bottom), mapped to the ground truth class with the highest overlap (best viewed in color). }
    \label{fig:latent_space}
\end{figure*}

The latent spaces of the best-performing models are shown in Fig.~\ref{fig:latent_space}. In the top row, we depict the ground truth and the predictions on the bottom. We notice qualitatively that the majority of the ground truth classes are separated in the latent space, and the unsupervised clustering prediction matches them. The models can learn representations for data points of the same driving event that can be separated by the $k$-means algorithm at a timestamp level. For T-Loss and TNC, we can observe that the `Turn Right' event has no matching cluster in the prediction (compare dashed circles), which is an indication why their macro $F_{1}$ score in Table~\ref{tab:metric_results} is lower than for the generative models.

\subsection{Segmentation}

\begin{figure}[t]
    \centering
    \includegraphics[width=0.8\linewidth]{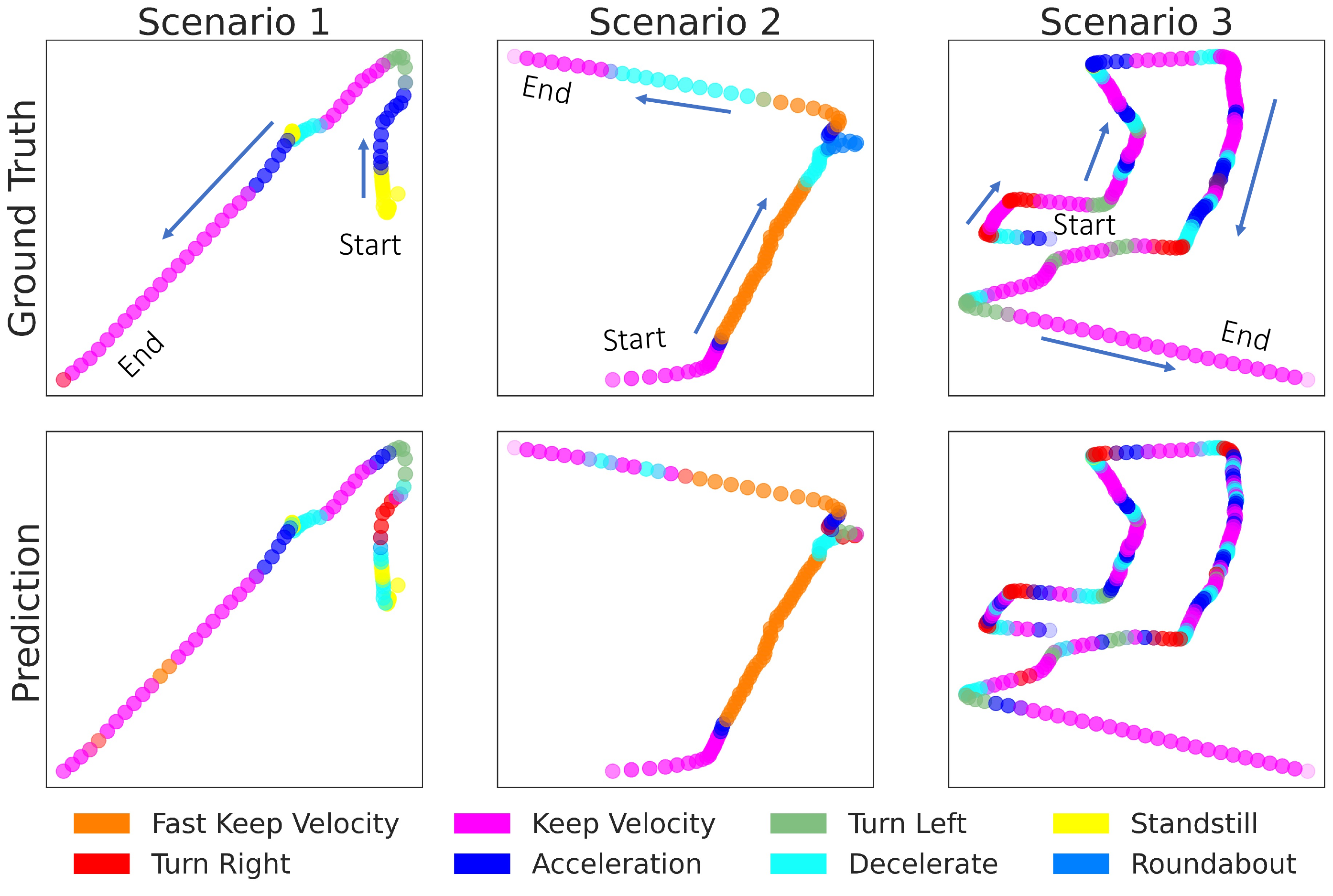}
    \caption{Segmentation results under different driving scenarios. Ground truth (top), and clustering (bottom), predicted clusters are mapped to the ground truth class with the highest overlap}
    \label{fig:segmentation}
\end{figure}

In Fig.~\ref{fig:segmentation}, we depict GPS trajectories of driving scenarios colored with a ground truth annotation (top row) and the segmentation (bottom row) of our best model w.r.t. Table~\ref{tab:metric_results}.
In Scenario~1 (left), the vehicle exits a parking lot. In Scenario~2 (middle), it drives a secondary road with a roundabout and enters a village. In Scenario~3 (right), it experiences a common stop-and-go scenario in the neighborhood of a city. 

We observe a high similarity in our predictions compared to the ground truth. In Scenario~1, we notice that the first right turn of the vehicle was annotated as an acceleration, but our model predicts a right turn. Because both events are correct, our model can support a human annotator. In Scenario~2, our model cannot capture the roundabout in one cluster, but it predicts corresponding turns. For Scenario~3, the approach is accurate in predicting the driving event states of the car. We notice that our prediction is sometimes more accurate than our annotation, which shows a) that an annotation process can be ambiguous and b) that our approach can support a human annotator during that process.

\subsection{Classification}
\begin{table*}[t]
    \centering
    \caption{Supervised macro $F_{1}$ classification results for each individual class on test trip, G=Generative, C=Contrastive}
    \begin{tabular}{c||p{0.05\linewidth}|p{0.04\linewidth}|p{0.04\linewidth}|p{0.03\linewidth}|p{0.03\linewidth}|p{0.05\linewidth}|p{0.03\linewidth}|p{0.03\linewidth}|p{0.045\linewidth}||p{0.025\linewidth}|p{0.04\linewidth}|p{0.04\linewidth}|p{0.03\linewidth}||p{0.03\linewidth}}
        & Standstill & Accele-ration & 
        Decele-ration & KV & 
        Fast KV & 
        Autonom KV & 
        Turn Left & 
        Turn Right & 
        Reverse & 
        S Turn & 
        Lane Change & 
       Round-about & Brake & Macro $F_{1}$\\
       \hline
        AE (G)& \textbf{0.839} & 0.359 & 0.630 & 0.463 & 0.537 & 0.188 & 0.341 & 0.222 & 0.322 & - & \textbf{0.014} & - & \textbf{0.025} & 0.337 \\ 
        \hline
        VAME* (G)& 0.749 & 0.552 & 0.672 & 0.719 & 0.705 & 0.218 &  0.345 & \textbf{0.223} & \textbf{0.662} & - & - & \textbf{0.047} & 0.010 & 0.416 \\
        \hline
        VAME (G)& 0.764 & \textbf{0.593} & \textbf{0.683} & 0.715 & 0.679 & \textbf{0.324} & \textbf{0.387} & 0.161 & 0.446 & - & - & 0.026 & - & 0.417\\
        \hline
        Drive2Vec (G) & 0.710 & 0.587 & 0.669 & \textbf{0.731} & 0.671 & 0.280 & 0.348 &  0.154 & 0.458 & - & - & 0.045 & - & 0.412\\
        \hline
        \hline
        T-Loss (C) & 0.742 & 0.471 & 0.652 & 0.676 & \textbf{0.717} & 0.204 & 0.320 & 0.035 & - & - & - & - & - & \textbf{0.438} \\
         \hline
        TNC (C) & 0.801 & 0.535 & 0.666 & 0.679 & 0.447 & - & 0.251 & - & 0.017 & - & - & - & - & 0.386\\
    \end{tabular}

    \label{tab:classification}
\end{table*}
We train a linear SVM with 10-fold cross-validation to evaluate the linear separability of the representations and the results are shown in Table~\ref{tab:classification}. The macro $F_{1}$ results show that both generative and contrastive models achieve similar performance, with the T-Loss model having a slightly better score on average. T-Loss achieves comparable results for most events when compared to the SOTA method Drive2Vec. TNC is outperformed by T-Loss due to its neighborhood estimation approach that is not well designed for CAN-data. 

The events `Standstill', `Acceleration', `Deceleration', `KV', and `Fast KV' are distinguished the best in the latent space. This result also reflects the clustering predictions of the latent space in Fig.~\ref{fig:latent_space}. For `Autonom KV', `Turn Left', `Turn Right' and `Reverse', we see a drop in the scores, which shows that the models perform worse for these events. The `Brake' events are not found because they are encoded similarly to `Deceleration' due to an electric vehicle's recuperation mechanism (compare signal BrkPrs\_Fl in Fig.~\ref{fig:example_cluster_segments}). Furthermore, the events that are composed of other events like `Roundabout' can not be discovered with our approach because the window size is too small and our approach finds the events that compose it.



\subsection{Simultaneous Clustering and Segmentation}

\begin{figure}  [t]
\centering
 \includegraphics[width=\linewidth]{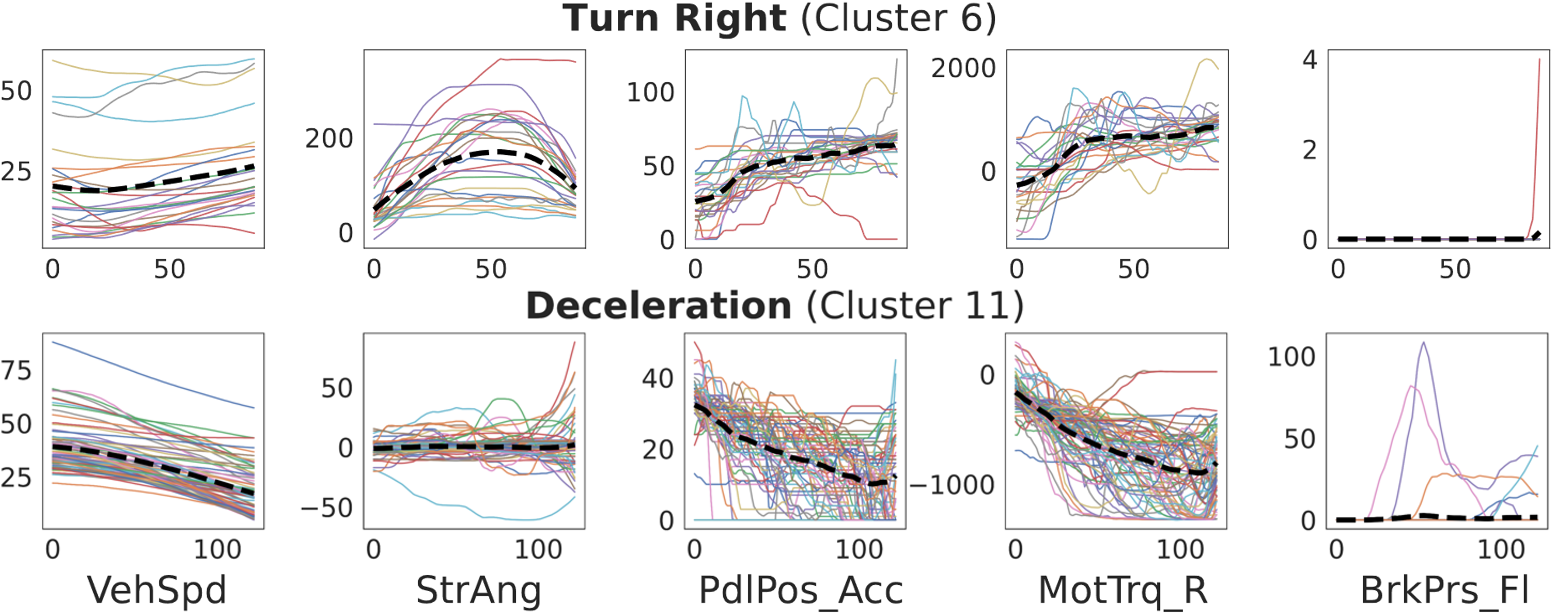}
 \caption{Exemplary clusters and their segments with real-world interpretation}
 \label{fig:example_cluster_segments}
\end{figure}

We verify the segmentation ability of our approach, in Fig.~\ref{fig:example_cluster_segments}, with all segments of two arbitrary clusters. They are visualized with five different vehicle CAN-data signals to interpret the events. From left to right, these channels are `Velocity', `Steering~wheel~angle', `Accelerator~pedal~position', `Motor~torque', and `Brake~pressure'.
We only show events longer than three seconds, because driving events usually last four to six seconds~\cite{zhang2014study}. Further, each segment is stretched to the length of the longest segment in its cluster. The dashed black line is the average of all segments. 

Each segment in the `Deceleration' cluster in Fig.~\ref{fig:example_cluster_segments} follows a similar pattern: The vehicle's velocity decreases, motor torque is at a negative value, and sometimes the brake was pressed. Accordingly, in the `Turn Right' cluster, the steering angle increases up to 200 degrees, which signals a clear right turn. The similar segments in both clusters verify that our approach can simultaneously cluster and segment vehicle CAN-data in an unsupervised manner.

\section{Conclusion \& Future Work}
We present an unsupervised approach that discovers driving events in vehicle CAN-data. Using self-supervised learning (SSL), we encode all time steps of driving events from the same category to a similar representation. 
This property allows clustering each time step in isolation, which simultaneously clusters and segments vehicle CAN-data into driving events. Our experiments show the effectiveness of our approach in unlabelled CAN-data from a Tesla Model 3 vehicle. Further, we find that representations from contrastive SSL methods perform comparably to state-of-the-art generative SSL methods for CAN-data.

Possible extensions of our approach include discovering different driver's behavior patterns for events or sequences of events with, for instance, a language model. In addition, we can improve our approach with better representation learning methods. For instance, we consider developing a better neighborhood estimation method for TNC that better captures the non-stationary nature of events in CAN-data. Furthermore, we believe that our approach is generally applicable to all kinds of multivariate time series data composed of sequences of discrete events. We intend to evaluate the approach for vehicles with a combustion engine or even in other domains (e.g., HAR, EEG) in the future. 


\addtolength{\textheight}{-12cm}   





\section*{ACKNOWLEDGMENT}
This work has been funded by the LOEWE initiative (Hesse, Germany) within the emergenCITY centre.


\bibliographystyle{IEEEtran}
\bibliography{IEEEabrv,mybibfile}

\end{document}